\documentclass[runningheads]{llncs}
\usepackage[T1]{fontenc}

\usepackage[marginal]{footmisc}

\usepackage[backref]{hyperref}
\hypersetup{
	hidelinks}
\usepackage{lineno,hyperref}
\usepackage{amsmath,amssymb,amsfonts}
\usepackage{algorithmic}
\usepackage{graphicx}
\usepackage{epstopdf}
\usepackage{textcomp}
\usepackage{xcolor}
\usepackage{subfigure}
\usepackage{multirow}
\usepackage{url}
\usepackage{makecell}

\begin{document}
\begin{sloppypar}
\title{Domain Knowledge Based Temporal-spatial Graph Convolution Network for ECG Recognition}
\titlerunning{Temporal-spatial Graph Convolution Network for ECG Recognition}
%
\author{Wenting Ma\inst{1,*}\orcidID{0000-0002-1245-9885} \and
 Zhipeng Zhang\inst{1,2,*} \and
Xiaohang Yuan\inst{2} \and
Ningwei Xie\inst{1} \and
Yuxin Xie\inst{3} \and
Xiaolin Wang\inst{2} \and
Meng Guo\inst{1} \and
Xingang Chai\inst{1} \and
Zhenjie Yao\inst{4,5}\orcidID{0000-0003-1027-637X}
}
\authorrunning{Z. Zhang et al.}
%
\institute{China Mobile Research Institute, Beijing, China \and
China Mobile GBA (Greater Bay Area) Innovation Institute, Guangzhou, China \and
Beijing Jiaotong University, Beijing, China  \and Institute of Microelectronics, Chinese Academy of Sciences, Beijing, China \and
Purple Mountain Laboratory, Nanjing, China \\
\email{yaozhenjie@ime.ac.cn} 
}
\maketitle              
\footnote{$^{*}$ Ma and Zhang contribute equally to this work.\\}

\begin{abstract}

In light of strides in Artificial Intelligence (AI) and its wide-spread application, challenges persist in the interpretability of AI models, particularly within specialized domains like healthcare, such as electrocardiograph (ECG) recognition. Rather than relying solely on end-to-end convolutional neural networks, this paper introduces a novel approach using a domain knowledge-based graph convolution network for ECG recognition. Key landmarks points of PRQST, vital to ECG interpretation, are incorporated as domain knowledge. The double-stream directed graph is employed to model both intra and inter ECG cycles. Specifically, spatial directed graphs capture the positional relationships among key points, while temporal directed graphs delineate temporal dependencies between adjacent cycles in extended ECG sequences. Experimental results on the First Chinese ECG Intelligent Competition dataset, which specifically classify ECG into nine categories, prove the efficacy of the proposed model. The overall average F1 score is 88.1\%, the average F1 score of rare categories is 76.3\%, both outperform the state-of-the-art models. The introduction of domain knowledge did enhance the detection performance, especially for rare categories.

\end{abstract}
\section{Introduction}
Recent advancements in artificial intelligence (AI) have led to its extensive adoption in various domains \cite{perera2014sensing,khan2020industrial}. This adoption is being explored to enhance the quality and reduce the cost of healthcare \cite{muhammad2020deep}. Clinicians stand to gain from computer-aided treatment, as it helps mitigate the impact of subjective interpretations and discrepancies among observers during the diagnostic process \cite{Yao2018Applying}.

Deep learning (DL) offers a promising approach to address these requirements \cite{Yao2018Applying}. However, the notable improvement in performance often comes from the large dataset. In traditional computer vision tasks, extensive and well-annotated datasets like ImageNet (with over 14 million labeled images from 20,000 categories) are available \cite{deng2009imagenet}. In healthcare applications, there are some rare cases where it is difficult to obtain corresponding data and annotations. There is a growing suggestion that incorporating domain knowledge into machine learning models could prove helpful in many situations \cite{von2021informed}. 

The domain knowledge of experts encompasses the specific regions they frequently focus on, the characteristics they prioritize, and their extensive anatomical knowledge. The domain knowledge is accumulated, summarized, and validated based on a vast number of samples. Despite the wealth of domain knowledge available, effectively integrating it into deep neural networks remains a challenging and as-yet-unresolved problem.
\begin{figure}[!t]%
	\centering
    \includegraphics[width=0.75\textwidth]{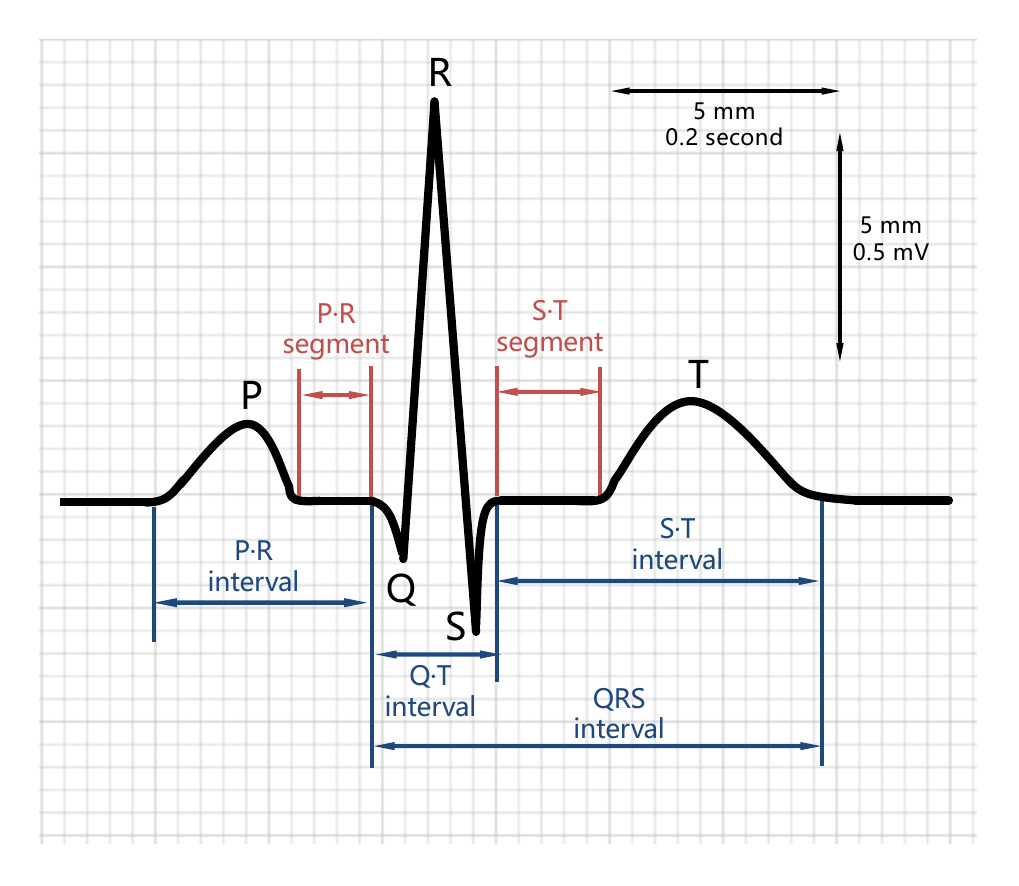}
	\caption{The composition of each wave cycle of electrocardiogram, including P wave, PR interval, QRS wave group, ST segment, T wave, QT interval, etc.}
    \label{fig:seg}
\end{figure}

The electrocardiogram (ECG) is widely utilized to identify a diverse range of cardiac issues \cite{carrara2015heart,kiranyaz2015real}. ECG interpretation is pivotal in assessing cardiovascular diseases, relying on the expertise of cardiologists or computer-aided diagnosis systems. Deep Convolutional Neural Networks (CNN) extract features and recognize patterns directly from raw ECG data in an end-to-end manner \cite{hannun2019cardiologist}. We posit that the integration of ECG domain knowledge into a deep neural network architecture can further enhance the diagnostic performance of ECG-based assessments. Figure \ref{fig:seg} illustrates the composition of each waveband of the electrocardiogram, which is the one cycle of the quasi-periodicity ECG signal. ECG abnormalities typically manifest as waveform abnormalities and rhythm abnormalities, including abnormalities within one cycle (intra-cycle abnormalities) and across long-term cycles (inter-cycle abnormalities). The information contained in the wavebands represents crucial domain knowledge for ECG diagnosis.

To capture the crucial domain knowledge in ECG signal, graph convolutional networks (GCN), which offers greater flexibility in data representation, is adopted to model both the time and voltage information associated with key points in the ECG. GCNs conceptualize signals as nodes within a topological network and capture their relationships through the edges of the graph. GCNs prove efficient in discovering and modeling the inherent connections between graph nodes \cite{dash2022inclusion}.

Our main contributions are as follows:
\begin{itemize}
	\item A domain knowledge-based graph structure is built for ECG, which represents the landmarks of the ECG and establish connections among them to effectively capture rhythm and morphology of an ECG.
	\item A double-stream directed graph model, which involves both spatial directed graphs (SDGs) for intra-cycle and temporal directed graphs (TDGs) for inter-cycle information, is applied to model the graph structure constructed from the ECG.
	\item Validating the effectiveness of the method on a real-world ECG dataset, especially for the detection of rare abnormal categories.
\end{itemize}

The structure of the paper is outlined as follows: Section \ref{sec-rw} provides a review of related works. The proposed model is detailed in Section \ref{sec-md}. Experimental results are shown and discussed in Section \ref{sec:exp}. Finally, the conclusion is drawn in Section \ref{sec-con}.
\section{Related Works}\label{sec-rw}
\subsection{Domain knowledge for Deep Learning}
Complex deep learning models require a large amount of labeled data for training, which not only limits the interpretability of the models but also makes it difficult to effectively model rare categories\cite{dash2022review,yin2019domain}. Integrating domain knowledge has become a feasible strategy to address this challenge. Domain knowledge can be summarized into the following categories: (1) Input data utilized by the deep network model, where the domain knowledge is represented in symbolic form; (2) Altering the loss function, which includes the penalty terms reflecting the incorporated domain-knowledge constraints. (3) Adapting the structure or parameters of the model, where domain knowledge can be integrated by imposing constraints on model parameters or by designing specific model structures. For example, the training progress of medical students can be incorporated into the deep neural network training process through curriculum learning \cite{raghu2019transfusion,ravishankar2016understanding}, which dictates a sequence of training examples that become progressively more difficult as the learner advances.
\subsection{ECG Classification}\label{subsec2.2}
Each heartbeat within the cardiac cycle illustrates the temporal evolution of the heart's electrical activity, characterized by distinct patterns of electrical depolarization and repolarization. Expert cardiologists can readily detect abnormalities in the heartbeat by identifying anomalies in the rhythm or changes in the morphological pattern of recorded ECG waveforms. However, automating this task proves to be a formidable challenge for computer systems, leveraging signal processing techniques such as frequency analysis, wavelet transform, filter banks, heuristic approaches, and hidden Markov models \cite{li1995detection}. Despite these efforts, the inherent variability in ECG signals among patients poses a significant obstacle. Several studies utilizing DNNs for the classification of ECG data have demonstrated notable success. For instance, Yao \cite{yao2017atrial} employed multi-scale convolutional neural networks to detect Atrial fibrillation, while Yuan et al. \cite{yuan2019diagnosing} proposed a deep neural network architecture enhanced with complex hand-crafted features. Additionally, attention-based models, such as the Res-BiLSTM-Net \cite{res2019arrhythmia} and a novel structure combining a deep residual network with an attention mechanism \cite{liu2019automatic}, have also yielded remarkable results. In the work by Kiranyaz et al. \cite{kiranyaz2015real}, 1-D Convolutional Neural Networks (CNNs) were employed to construct a fast and accurate ECG classification and monitoring system. Peng et al. \cite{peng2024deep} constructed a deep learning framework that can simultaneously perform denoising and classification.
\subsection{Graph Convolutional Neural networks}\label{subsec2.3}
Graph convolutional neural networks are widely used in various fields \cite{georgousis2021graph}, such as molecular structures \cite{na2020scale}, transportation networks \cite{cui2019traffic}, backbone networks \cite{yao2021internet}, social networks \cite{wang2021community}. It is worth noting that, the skeleton-based approach for human action recognition has gained significant attention due to its robustness against changes in camera viewpoints, variations in human body scales, and interference from backgrounds \cite{feng2022skeleton}. GCN have played an increasingly critical role in advancing skeleton-based human action recognition. Xie proposed a novel correlation-driven joint-bone fusion GCN for pose prediction \cite{xie2022attention}. Given the success of GCNs in tasks like human action recognition, there is a belief that GCNs can enhance the classification of ECG signals, demonstrating the potential for these graph-based neural networks across diverse applications.

This paper extracts the skeleton of the ECG signal as the input to a graph convolutional neural network, thereby embedding and representing domain knowledge. The graph information extracted from the key points of the ECG signal includes both morphological information of the waveform and rhythm, achieving a comprehensive modeling of the ECG signal.
\section{Method}\label{sec-md}
\begin{figure*}[!t]%
	\centering
	\includegraphics[width=1.0\textwidth]{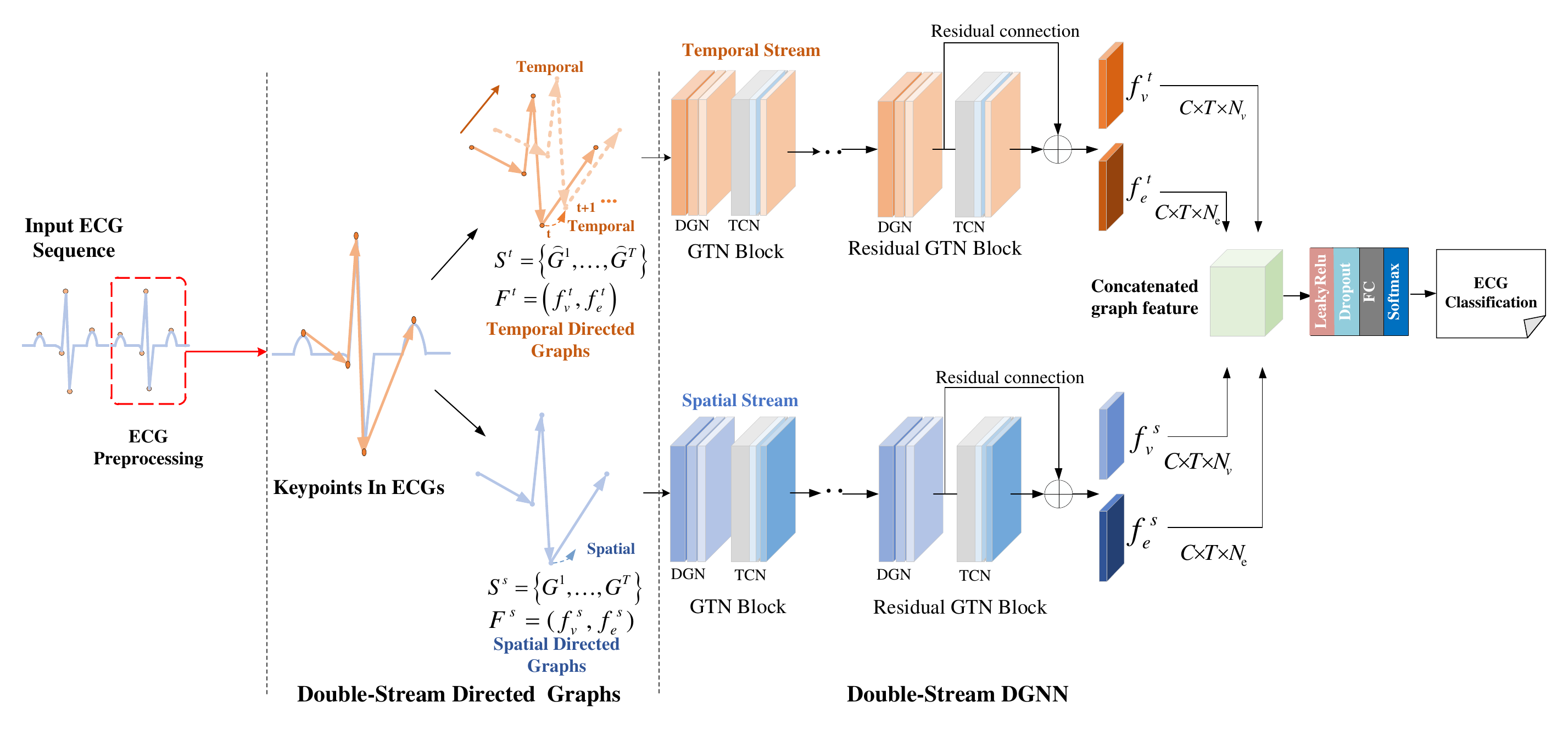}
	\caption{Architecture of the proposed method.}\label{fig:arc}
\end{figure*}

In this section, we present the proposed neural network model for ECG analysis, beginning with the preprocessing of ECG data. Subsequently, we provide a detailed explanation of the double-stream directed graph model. The architecture of the proposed neural networks is illustrated in Figure \ref{fig:arc}.
\subsection{ECG Preprocessing}\label{subsec3.1}
The ECG signal undergoes initial preprocessing through an analog band-pass filter with a cutoff frequency of 50 Hz. Following this, an analog-to-digital converter (ADC) samples the ECG signal at a rate of 500 Hz. Subsequently, the digitized signal undergoes a series of processing stages, which include the implementation of three linear digital filters. The first filter comprises cascaded low-pass and high-pass filters with integer coefficients, aiming to suppress noise from diverse sources such as muscle artifacts, electrode mobility, power-line interference, baseline drift, and high-frequency properties in T waves that may be comparable to those of QRS complexes. The application of digital filters is integral to attenuate the impact of various noise sources, thereby enhancing the signal-to-noise ratio.

To further enhance the quality of the QRS complexes, the signal undergoes a nonlinear transformation, such as differentiation or the square root operation. Automatic ECG signal analysis heavily relies on the accurate identification of QRS complexes. Once identified, further analyses of the ECG signal, such as heart rate and ST segment examination, can be conducted. The identification of QRS complexes is a challenging task due to both the physiological variability of these complexes and the diverse forms of noise present in ECG data.

Key points of the electrocardiogram are extracted using a dyadic wavelet transform. The identification of significant fluctuations in the ECG signal is achieved by locating local maxima in the wavelet transform modulus at various scales. This approach facilitates the extraction of key points in the signal, serving as a foundation for subsequent analysis. For a comprehensive understanding of this methodology, please refer to \cite{pan1985real}.
\subsection{Double-Stream Directed Graph Feature}\label{subsec:dsdg}
Double-Stream Directed Graph Feature is a feature extraction method for image recognition and computer vision, which is built on the basis of directed graph, where each node represents a local region in an image and each directed edge represents the spatial relationship between two nodes \cite{xie2021sequential}.

In this approach, each stream represents a perspective or a feature representation. For example, the first stream can extract color information, while the second stream can extract shape information. The information from these graph streams is combined into a directed graph and the edge weights between the nodes in the graph are calculated. Finally, classification or identification is made based on the characteristics of these nodes and edges. This method can improve the accuracy and has good universality for different types of data.

The heart generates electrical signals by alternating contraction and relaxation through the self-motivating function of cardio myocytes. A complete cycle of ECG signal waveform includes P wave, QRS wave and T wave. When using ECG to diagnose heart diseases, doctors will make the final judgment according to the amplitude, duration and other indicators of P, Q, R, S and T. In order to assist doctors in diagnosis, a directed acyclic graph is constructed to represent the waveform of the ECG signal, and use the spatial-temporal variation of the directed acyclic graph to represent the changes of ECG signal.

\begin{figure}[t]%
	\centering
	\includegraphics[width=1.0\textwidth]{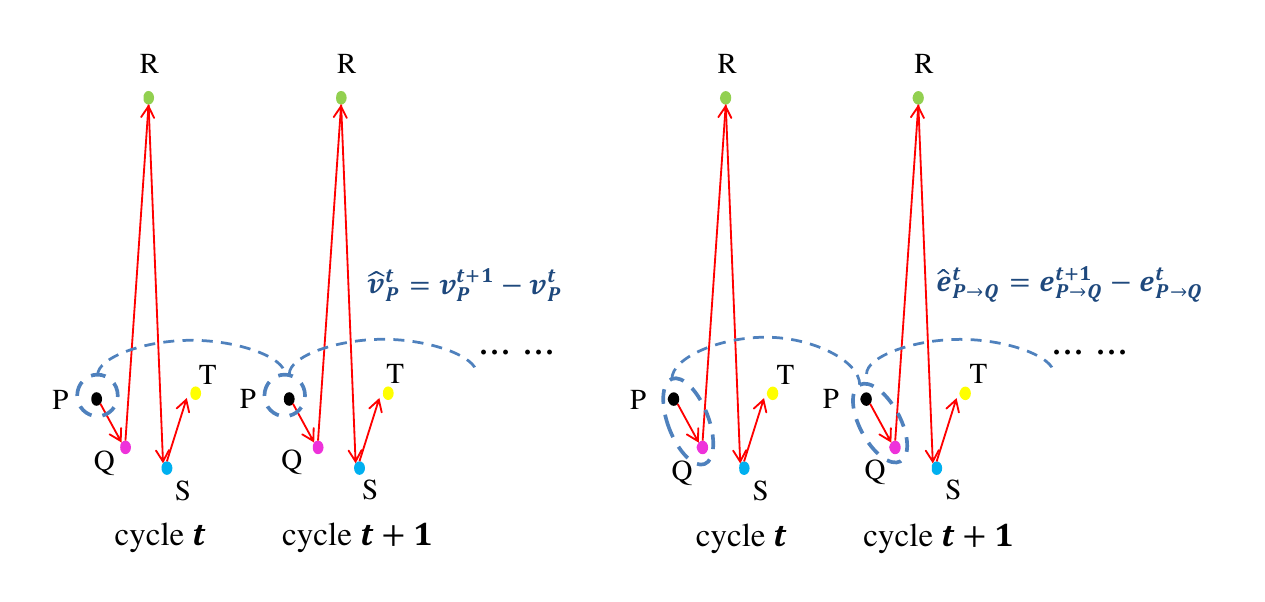}
	\caption{The key point displacement is defined by the temporal difference of coordinate vectors in two adjacent cycles. We display examples of vertex P and edge P to Q.}\label{fig:con}
\end{figure}

For ease of expression, we first define graph and its related concepts:

\textbf{\textit{Vertex}}: A point in a graph.

\textbf{\textit{Edge}}: An Edge between a vertex $V_{i}$ and $V_{j}$ is an Undirected edge and is represented by an undirected pair $e\left( V_{i},V_{j} \right) $.

\textbf{\textit{Graph}}: A graph $G$ consists of a finite set of vertices $V$ and a multiset of edges $E$, denoted here as $G=V\cup{E}$.

\textbf{\textit{Arc}}: If the edge from the vertex $V_{i}$ to $V_{j}$ has a direction, then this edge is called a directed edge, also called Arc.

\textbf{\textit{Diagraph}}: A diagraph $D$ is represented by a multiset of vertices $V$ together with arcs $A$, written as $D=V\cup{A}$.

\textbf{\textit{Ring}}: We define open trial as a path where no vertices are traversed more than once (all vertices are different). In the path $\varepsilon\left( V_{0},V_{n} \right) $, the vertex $\left\lbrace V_{1},\ldotp\ldotp\ldotp,V_{n-1} \right\rbrace $ is called the inner vertex of $\varepsilon\left( V_{0},V_{n} \right) $. If all vertices are different except $V_{0}=V_{n}$ and it contains at least one edge, then we call it a ring, which is the definition of a closed trail.

\textbf{\textit{Directed acyclic graph}}: A graph $G$ or a digraph $D$ that contains no cycle is called an acyclic graph.

Figure \ref{fig:con} illustrates how to build the connections of intra and inter cycle, which constructs a directed acyclic graph from electrocardiogram. The intra cycle connection is from the previous key point to the first key point following it, which results in $P^t\rightarrow Q^t\rightarrow R^t\rightarrow S^t\rightarrow T^t$. The inter cycle connections are one to one. For example, the connections between $t$ and $t+1$ cycles include: $P^t\rightarrow P^{t+1}$, $Q^t\rightarrow Q^{t+1}$, $R^t\rightarrow R^{t+1}$, $S^t\rightarrow S^{t+1}$, $T^t\rightarrow T^{t+1}$.

Thus, in an electrocardiogram cycle, the key points and their connections can match the vertices and directed edges of the acyclic graph. We use the graph $G=\left( V,E \right) $ to represent ECG.

\begin{equation}
	V = \left\lbrace v_{m} \right\rbrace,m=1,2,\ldotp\ldotp\ldotp,N_{v}
\end{equation}
\begin{equation}
	E = \left\lbrace e_{k}=v_{k+1}-v_{k} \right\rbrace,k=1,2,\ldotp\ldotp\ldotp,N_{e}
\end{equation}
\begin{equation}
	M = \left\lbrace M_{m,k} \right\rbrace,m=1,2,\ldotp\ldotp\ldotp,N_{v} \, and \, k=1,2,\ldotp\ldotp\ldotp,N_{e}
\end{equation}
where $V$ represents the set of vertices, $E$ represents the set of directed edges, $N_v$ is the number of nodes, $N_e$ is the number of directed edges. The directed edges always point from the previous key point to the next key point. The initial graph topology is represented by a matrix $M$.

We then extract the spatial and temporal characteristics of the vertices and edges of the directed acyclic graph from the ECG. $s$ and $t$ indicates spatial and temporal, respectively. We use $G^{t}$ to represent the space digraph of cycle $t$. Spatial flow consists of a series of spatial directed graphs, using $S^{s}=\left\lbrace G^{1},\ldotp\ldotp\ldotp,G^{T} \right\rbrace$. The vertices and edges of a spatial digraph are represented by $F^{s}=\left(f_{v}^{s},f_{e}^{s}\right)$, and the temporal digraph can be denoted by $F^{t}=\left(f_{v}^{t},f_{e}^{t}\right)$. The space vertex attribute $f_{v}^{s}$ and the temporal vertex attribute $f_{v}^{t}$ are key point displacement matrices with size $C_{in}\times T \times N_{v}$. The spatial edge attribute ${f_{e}^{s}}$ is composed of the changes of vertex in each cycle relative to the following one, and the temporal edge attribute ${f_{e}^{t}}$ is composed of the changes of each vertex in the adjacent cycle. ${f_{e}^{s}}$ and ${f_{e}^{t}}$ are matrices of size $C_{in}\times T \times N_{e}$. As shown in Figure \ref{fig:con}, the key point displacement is defined by the temporal difference of coordinate vectors between two adjacent cycles, i.e.,$\hat{V}_{k}^{t}=V_{k}^{t+1}-V_{k}^{t}$, where $V_{k}^{t}$ and $V_{k}^{t+1}$ are the coordinate vectors of $V_{k}$ at cycle $t$ and $t+1$, respectively.

We can use the attributes of vertex and edge to calculate the spatial transformation and temporal transformation between each connected vertexes. Specifically, we can calculate the time and amplitude difference of the vertex $v_{k}$ to its successor $v_{k+1}$. Or calculate the difference from the edge $e_{m}$ to $e_{m+1}$ from the $v_{k}$ and the successor vertex $v_{k+1}$, respectively.

\subsection{Double-Stream Directed Graph Framework}\label{subsec3.3}

Directed graph neural networks (DGNN) use the spatial and temporal characteristics of ECG signals for classification. As shown in Figure \ref{fig:arc}, the entire network consists of multiple layers, each of which has a graph containing vertex and edge attributes, and outputs the same graph with updated attributes. At each level, vertex and edge properties are updated based on adjacent edges and vertices. At the bottom level, each vertex or edge can only receive attributes from adjacent edges or vertices, and the model in these layers is designed to extract local information about vertices and edges when updating attributes. At the top level, information from key points that are can propagated to each other by accumulate multi-layer information aggregation. DGNN models the data structure of the tree, directed acyclic graph.

DGNN is composed of directed graph network (DGN) block \cite{li2017skeleton}, Each DGN contains two update function $h^{v},h^{e}$, and two aggregate function $g^{e^{-}},g^{e^{+}}$. The update function is used to update attributes of vertices and edges based on adjacent edges and vertices. The aggregation function is used to aggregate properties contained in multiple edges connected to a vertex. Finally, the spatial branch of DGNN outputs the encoded vertex feature ${f_{v}^{s}}$ and encoded edge feature ${f_{e}^{s}}$ \cite{xie2022attention}. Similarly, the temporal branch outputs the encoded vertex feature ${f_{v}^{t}}$ and encoded edge feature ${f_{e}^{t}}$.

For each cycle, we form a graph by concatenating the temporal and spatial graphs, where each node contains all encoded spatial-temporal features of its corresponding vertex and edge. Therefore, four encoded feature tensors of the DGNN outputs are stacked along channel-dimension to obtain the concatenated spatial-temporal graphs feature. In this way, the final feature encodes the spatial-temporal characteristics of ECG sequence. We use average-pooling to reduce dimensions of the result feature and use fully-connected to project it into a vector of size $1\times N_{cls}$. Finally, SoftMax function was adopted to make the prediction for the ECG labels.
\section{Experiments}\label{sec:exp}
This section begins with an introduction to the dataset, followed by the experimental setting, measurements, results and discussion.
\subsection{Dataset}\label{subsec4.1}

\begin{table}[htbp]
	\renewcommand\arraystretch{1.3}
	\setlength\tabcolsep{6pt}
	\caption{There are eight typical cardiovascular abnormalities in the dataset.}
	\begin{center}
		\begin{tabular}{|c|c|}
			\hline
			\textbf{Abbreviations} & \textbf{Categories} \\
			\hline
			AF & Atrial Fibrillation\\
			\hline
			TWC & T-wave Changes\\
			\hline
			ER & Early Repolarization\\
			\hline
			LAFB & Left Anterior Fascicular Block\\
			\hline
			CRBBB & Complete Right Bundle Branch Block\\
			\hline
			PVC & Premature Ventricular Contractions\\
			\hline
			PAC & Premature Atrial Contractions\\
			\hline
			FDAVB &  \makecell{First-degree Atrioventricular Heart \\ Block Diagnosing Cardiac Abnormalities}\\
			\hline
		\end{tabular}
		\label{tab2}
	\end{center}
\end{table}

To validate our approach, we conducted a series of experiments on the datasets from the First China ECG Intelligent competition \cite{ekgdata2019}. The ECG data were sampled at 500 Hz and captured using the standard 12-lead format, encompassing six limb leads (I, II, III, aVL, aVR, and aVF) and six precordial leads (V1, V2, V3, V4, V5, and V6). These ECGs' length ranges from 4,500 to 30,000, corresponding to a recording time of 9.5s to 60s. There are 6,500 ECGs in the dataset, trained and tested through 5-fold cross validation. The dataset have eight typical cardiovascular abnormalities as shown in Table \ref{tab2}. It's worth noting that the labels for these abnormalities are not mutually exclusive; an ECG recording can possess multiple labels. The model should classify the ECG as normal if no abnormalities are detected.

\subsection{Evaluation Metric}\label{subsec4.2}

In this paper, we use the average $F_{1}$ score as the evaluation metric. To be specific, the average $F_{1}$ score is calculated by:
\begin{equation}	
	F_{1} = \frac{1}{9} \sum\limits_{i=0}^{8} F^{i}_{1},
\end{equation}
where $F^{i}_{1}$ is the $F_{1}$ score of the $i$th class, defined by
\begin{equation}	
	F^{i}_{1} = \frac{2P^{i} \times R^{i}} {P^{i}+R^{i}},
\end{equation}
in which \(P^{i}\) is the precision of the $i$th class, given by \(\frac{TP^{i}}{TP^{i} + FP^{i}}\) (true positives divided by the sum of true positives and false positives), \(R^{i}\) is the recall $i$th class, given by \(\frac{TP^{i}}{TP^{i} + FN^{i}}\) (true positives divided by the sum of true positives and false negatives), \(TP^{i}\) is the number of true positives, \(FP^{i}\) is the number of false positives, and \(FN^{i}\) is the number of false negatives.
\subsection{Implementation Details}\label{subsec4.3}
In the Double-Stream-DGNN, one GTN and two residual GTN blocks constitute each DGNN branch. Both the DGN and the TCN have a batch normalization (BN) layer and a Leaky ReLU layer behind them to make the network easier to converge. The dimensions of three blocks' output channels are set to 32, 64 and 64, respectively. The size of the temporal convolution kernel of three blocks are 3, 5, 7, and the stride of the last GTN is 2. We concatenated the output features extracted from the spatial-temporal branches of DGNN. To prevent over-fitting, we used the dropout layers with a rate of 0.2. Finally, we added a fully connected layer and a softmax layer at the end for class prediction. The Double-Stream-DGNN was trained to minimize the cross-entropy loss function, the stochastic gradient descent (SGD) optimizer, initial learning rate 0.001 and the batchsize is 32. The Double-Stream-DGNN is trained for 200 epochs.
\subsection{Results and discussion}\label{sec5}

\begin{table}[htbp]
	\renewcommand\arraystretch{1.3}
	\setlength\tabcolsep{6pt}
	\caption{$F_{1}$ score of the ECG abnormalities classification on the test set}
	\begin{center}
		\begin{tabular}{|c|c|c|c|c|c|}
			\hline
			\multirow{2}{*}{\textbf{Categories}}&\multirow{2}{*}{\textbf{Number}}&\multicolumn{4}{c|}{\textbf{Methods}} \\
			\cline{3-6}
			 & & \textit{Method1\cite{yuan2019diagnosing}}& \textit{Method2\cite{res2019arrhythmia}}& \textit{Method3\cite{liu2019automatic}} & \textit{   Ours   } \\
			\hline
			Normal & 1,953 & 0.914	& 0.896	 & 0.875 & \textbf{0.921}\\
			AF & 478 &	0.962&	\textbf{0.977}&	0.974&	0.958\\
			TWC& 2,049 &	\textbf{0.892}&	0.792&	0.757&	0.848\\
			ER& 222 &  0.500&	0.733&	0.736&	\textbf{0.745}\\
			LAFB& 164 &	\textbf{0.944}&	0.722&	0.747&	0.780\\
			CRBBB& 926 & \textbf{1.000}&	0.982&	0.983&	0.963\\
			PVC& 627 &	0.965&	0.963&	\textbf{0.971}&	0.965\\
			PAC& 628 &	0.874&	0.899&	0.926&	\textbf{0.938}\\
			FDAVB& 492 &	0.860&	\textbf{0.918}&	0.901&	0.813\\
			\hline
			Average& 838 &	0.879 &	0.876&	0.875&	\textbf{0.881}\\
			\hline
			LAFB \& ER& 164+222 & 0.722 & 0.728 & 0.742 & \textbf{0.763}\\
			\hline
			\multicolumn{5}{l}{*Bold marks 1st place in different categories.}
		\end{tabular}
		\label{tab1}
	\end{center}
\end{table}

In order to prove the superiority of the proposed method, we conducted a comparative study of previous state-of-the-arts methods on the ECG recognition, the detailed scores shown in Table 1. The compared methods include (1) deep convolutional neural networks enhanced with sophisticated hand crafted features \cite{yuan2019diagnosing}, (2) an attention-based Res-BiLSTM-Net model \cite{res2019arrhythmia} and (3) a structure that combines a deep residual network with an attention mechanism \cite{liu2019automatic}. Test results (F1 score) of all the four methods on the test set are shown in Table \ref{tab1}, the overall F1 score of the proposed classifier is 0.881, which achieves the highest score among the four models. For the Normal, premature ventricular contractions (PAC) and early repolarization (ER), the F1 score  increased to 0.921, 0.938 and 0.745, respectively. The proposed method exhibited commendable overall performance on the dataset.

In addition to overall performance, we pay particular attention to the performance of some rare abnormal categories. As we known, deep learning models are purely data-driven, and therefore, their performance often depends on the amount of data available. As shown in Table \ref{tab1}, the performance of LAFB and ER, which have the smallest data volume, is the poorest. The introduction of domain knowledge can alleviate the performance degradation caused by limited data volume. We separately calculated the average F1 scores of LAFB and ER, where the F1 score (0.763) of our model is significantly higher than other deep neural network models (0.722, 0.728, and 0.742) that do not incorporate domain knowledge. The introduction of domain knowledge can significantly improve the detection performance of rare abnormal categories. 

\begin{table}[htbp]
	\renewcommand\arraystretch{1.3}
	\setlength\tabcolsep{6pt}
	\caption{\textit{Precision}, \textit{Recall} and \textit{$F_{1}$ score} of our ECG abnormalities classification method on the test set}
	\begin{center}
		\begin{tabular}{|c|c|c|c|c|}
			\hline
			\multirow{2}{*}{\textbf{Categories}}&\multirow{2}{*}{\textbf{Number}}&\multicolumn{3}{c|}{\textbf{Metrics}} \\
			\cline{3-5}
			& & \textit{Precision}& \textit{Recall}& \textit{$F_{1}$ score} \\
			\hline
			Normal & 1,953 & 0.915 & 0.927 & 0.921\\
			AF     & 478   & 0.970 & 0.948 & 0.958\\
			TWC    & 2,049 & 0.886 & 0.813 & 0.848\\
			ER     & 222   & 0.794 & 0.702 & 0.745\\
			LAFB   & 164   & 0.812 & 0.750 & 0.780\\
			CRBBB  & 926   & 0.948 & 0.978 & 0.963\\
			PVC    & 627   & 0.959 & 0.971 & 0.965\\
			PAC    & 628   & 0.941 & 0.935 & 0.938\\
			FDAVB  & 492   & 0.857 & 0.773 & 0.813\\
			\hline
			Average& 838   & 0.884 & 0.869 & 0.881\\
			\hline
		\end{tabular}
		\label{tab3}
	\end{center}
\end{table}

From the Table \ref{tab3}, it is evident that despite our best efforts to mitigate the impact of data imbalance, the Recall for disease types with small samples (such as ER, LAFB and FDAVB) are also lower. This implies a higher likelihood of misclassification into other categories. Conversely, the Recall rate for disease types with large sample sizes is relatively higher compared to Precision. This observation aligns with our findings. Based on the above indicators, it can be concluded that the method proposed in this paper exhibits good stability.

\begin{figure*}[t]%
	\centering
	\hspace{-0.9cm}
	\subfigure[]
	{	
		\begin{minipage}[b]{.4\linewidth}
			\centering
			\includegraphics[width=120pt]{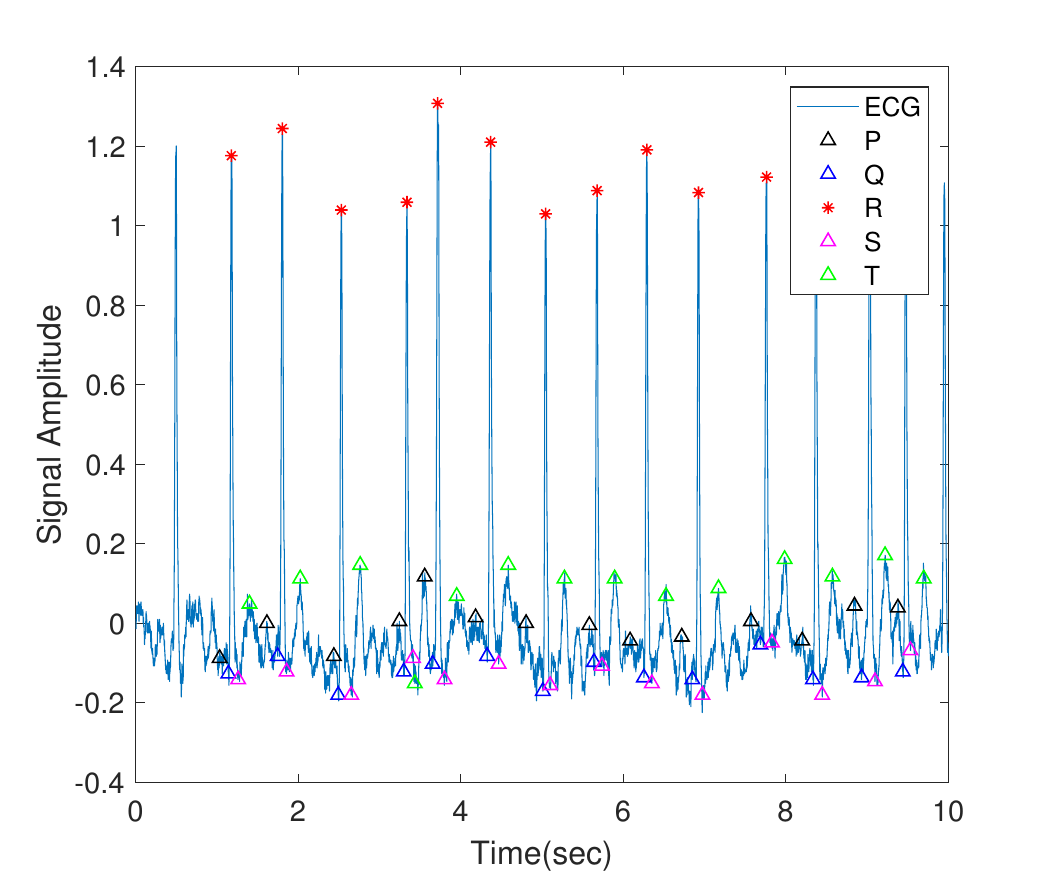}  
			\label{fig:exam1a}
		\end{minipage}
	}
	\hspace{-1.4cm}
	\subfigure[]
	{	
		\begin{minipage}[b]{.4\linewidth}
			\centering
			\includegraphics[width=120pt]{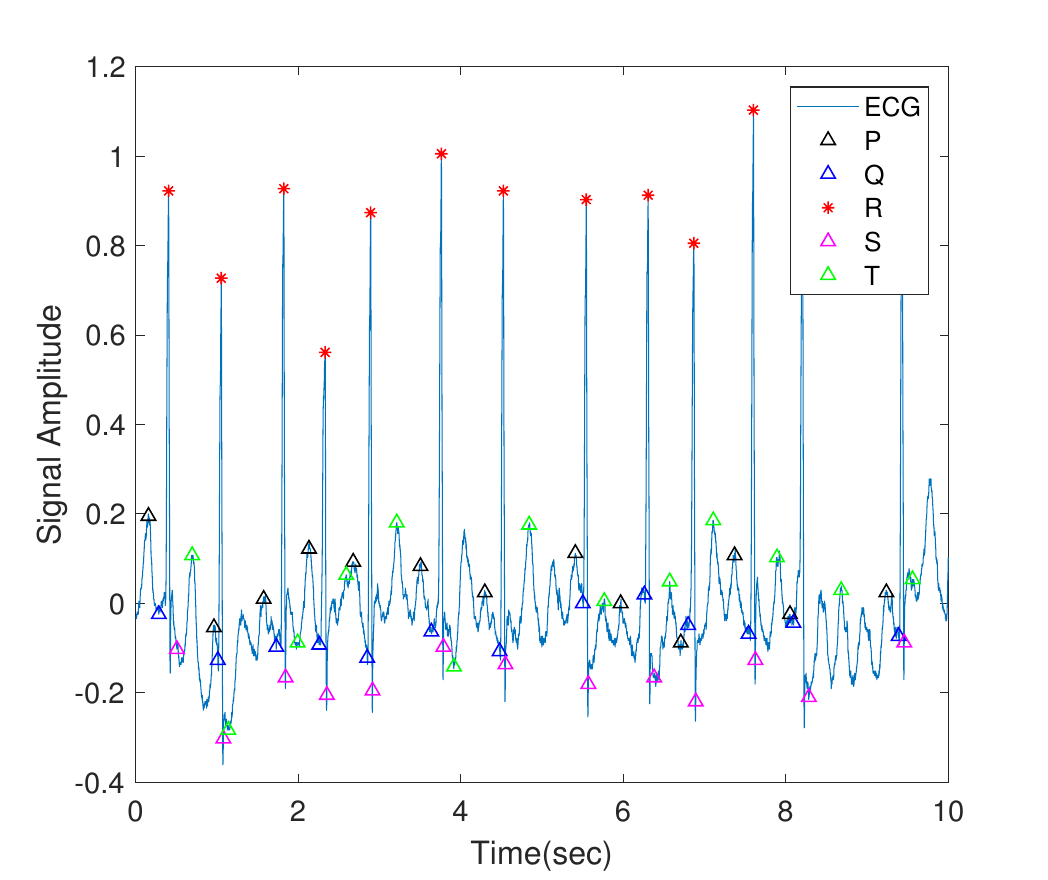}
			\label{fig:exam1b}
		\end{minipage}
	}
	\hspace{-1.4cm}
	\subfigure[]
	{
		\begin{minipage}[b]{.4\linewidth}
			\centering
			\includegraphics[width=120pt]{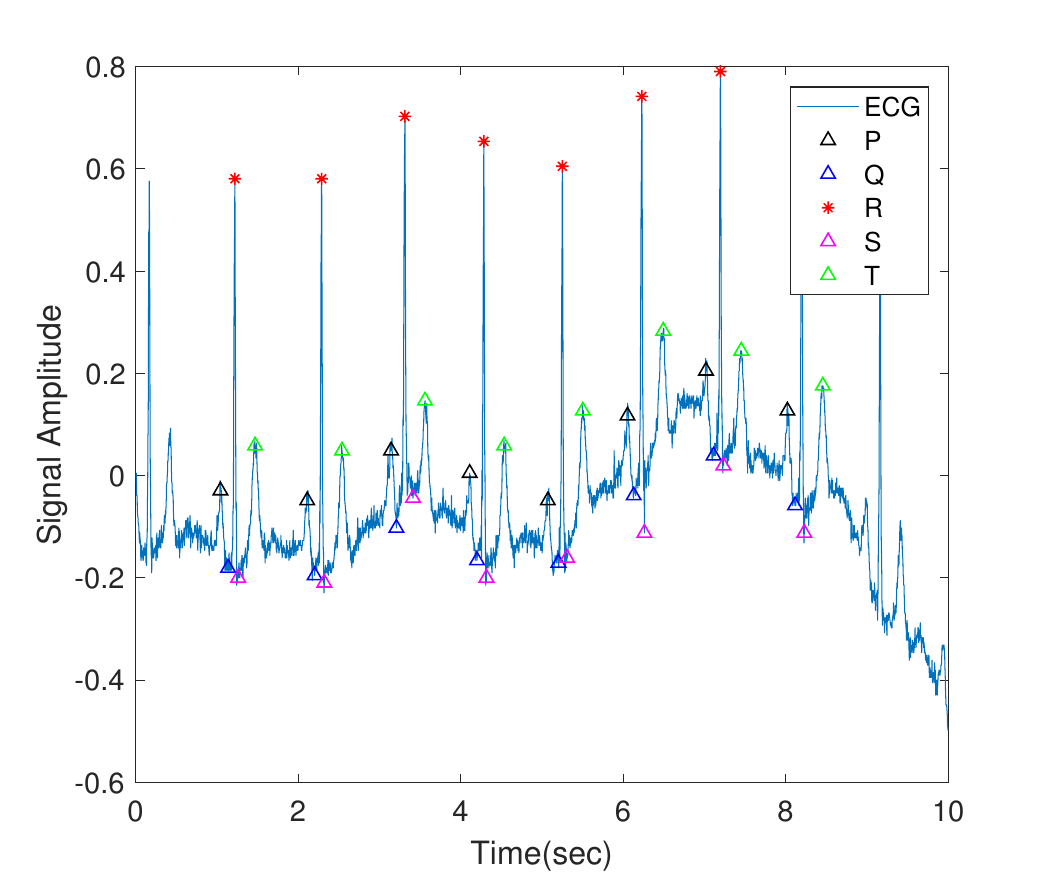}
			\label{fig:exam1c}
		\end{minipage}
	}
	\caption{(a) and (b) are the correctly classified ECG; (c) is a misclassified ECG (misidentification of key points due to interference).}\label{fig:exam1}
\end{figure*}

It is accepted that trust comes from understanding of how decisions are made, and what are the determinants of these decisions. One important advancement of domain knowledge fused method is that the models become more comprehensible. Domain-knowledge can be used in two different ways to assist this. First, it can constrain the kinds of models that are deemed understandable. Secondly, it can provide meaningful concepts to use in a model. Figure \ref{fig:exam1} displays 3 detected abnormal ECG results. Figure \ref{fig:exam1a} demonstrates that the method is able to detect not only anomaly in a single cycle (the last cycle on Fig. \ref{fig:exam1a}), but also the rhythm of the abnormal over several cycles (Figure \ref{fig:exam1b}). The performance of the detected abnormal ECG is straightforward to explain. The misclassified ECG is shown in Figure \ref{fig:exam1c}, The main reason for the failure is excessive noise, which leads to inaccurate detection of key points.

\begin{figure*}[t]%
	\centering
	\hspace{-0.9cm}
	\subfigure[]	
	{
		\begin{minipage}[b]{.4\linewidth}
			\centering
			\includegraphics[width=120pt]{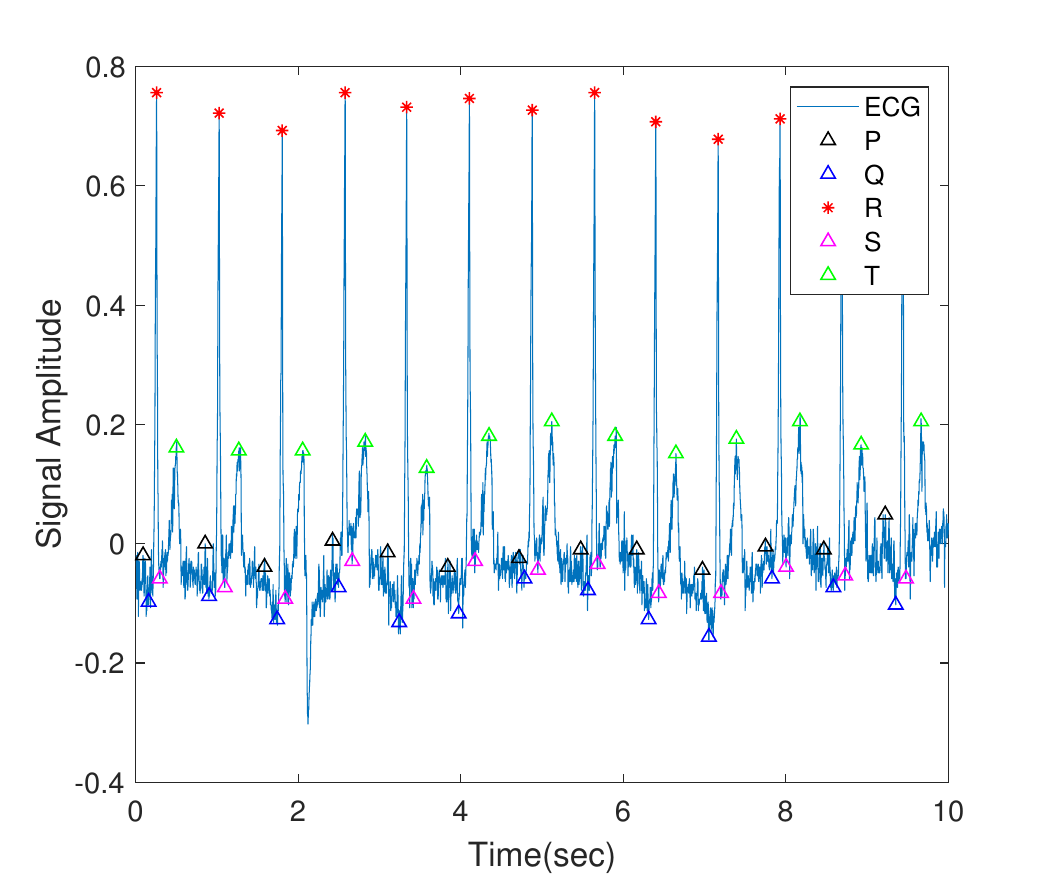}  
			\label{fig:exam2a}
		\end{minipage}
	}
	\hspace{-1.4cm}
	\subfigure[]
	{
		\begin{minipage}[b]{.4\linewidth}
			\centering
			\includegraphics[width=120pt]{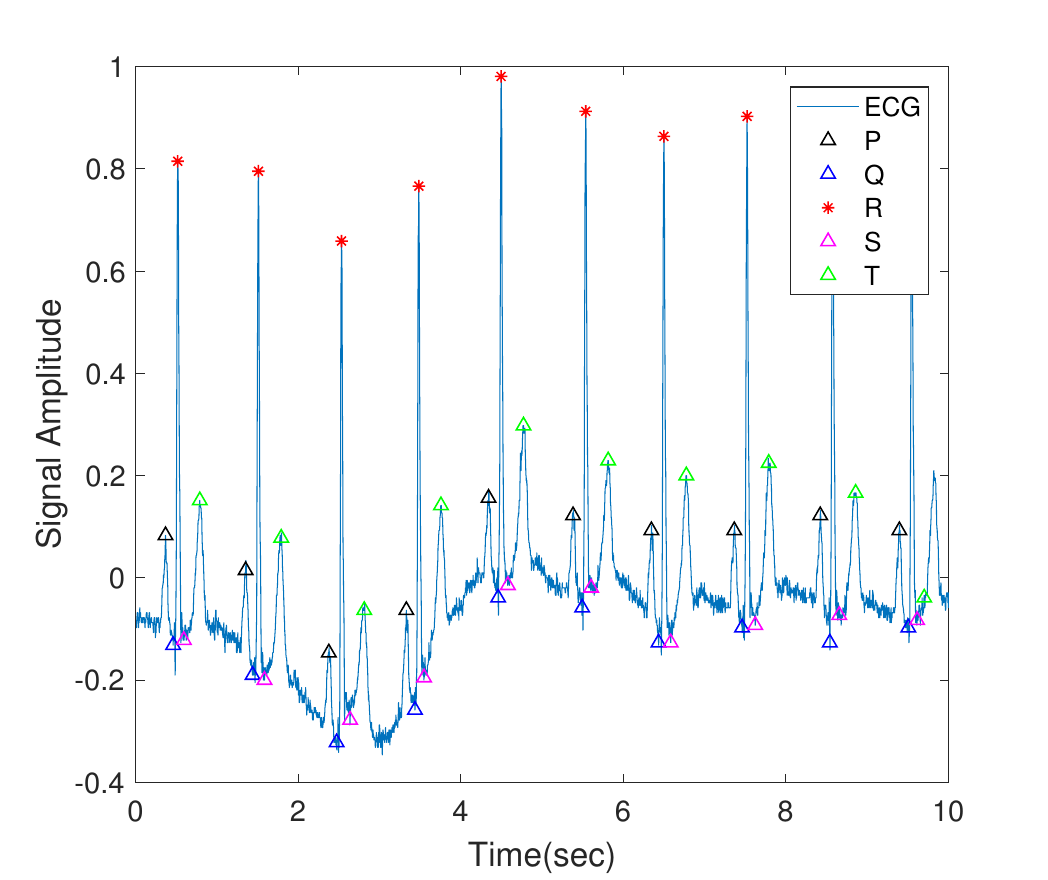}
			\label{fig:exam2b}
		\end{minipage}
	}
	\hspace{-1.4cm}
	\subfigure[]
	{
		\begin{minipage}[b]{.4\linewidth}
			\centering
			\includegraphics[width=120pt]{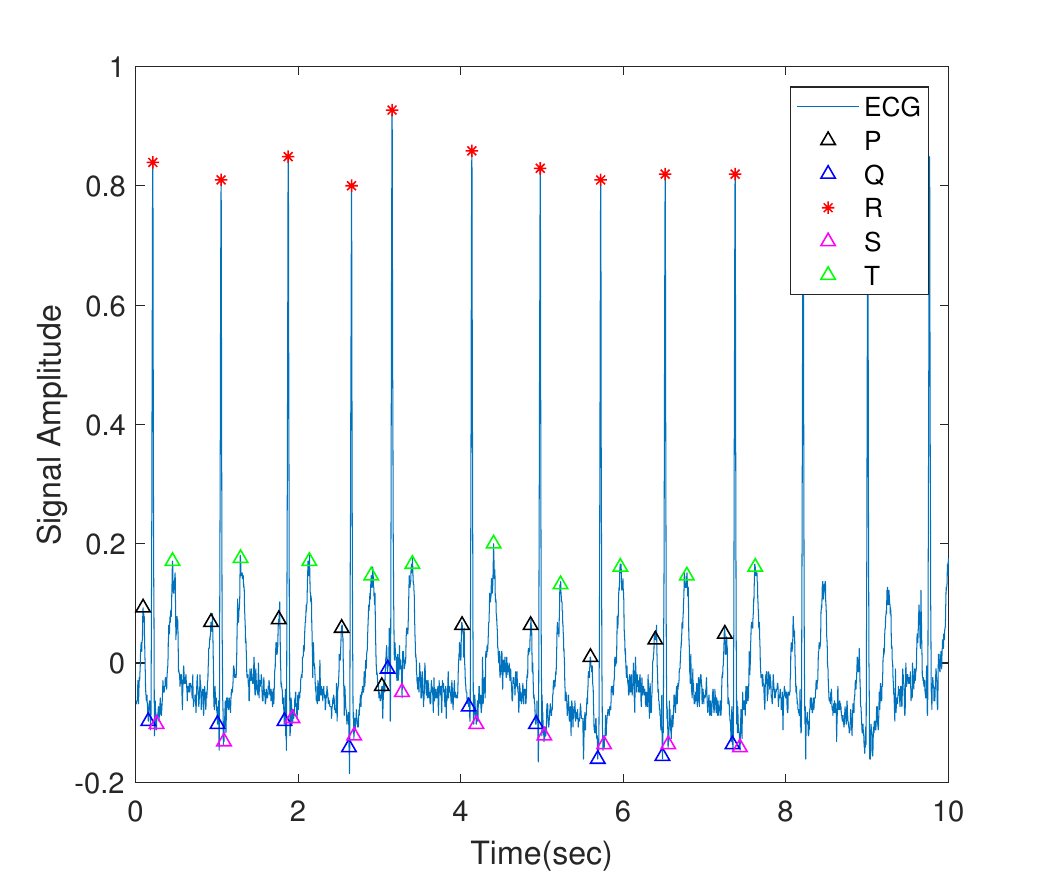}
			\label{fig:exam2c}
		\end{minipage}
	}
	\caption{(a) and (b) are the normal ECGs; (c) is an abnormal ECG with premature atrial contractions (PAC).}\label{fig:exam2}
\end{figure*}

Figure \ref{fig:exam2} illustrates the ability of our algorithm to accurately identify normal and abnormal ECGs. Specifically, Figure \ref{fig:exam2a} and Figure \ref{fig:exam2b} depict normal ECGs that exhibit noise and baseline drift, which would lead to misclassification with high probability using methods that focus solely on the whole ECG signal. However, our algorithm preprocesses the ECG signal and extracts key points, thereby minimizing the negative effects of noise and baseline drift. Furthermore, when dealing with abnormal ECG signals, our algorithm not only pays attention to waveforms but also to the temporal relationship between beats, resulting in a significant improvement in the classification accuracy of relatively rare anomalies. Figure \ref{fig:exam2c} shows an abnormal ECG with premature atrial contractions (PAC), which is the ECG signal of an atrial premature beat, and the anomaly occurred on the fifth heart beat only. As demonstrated in Table 1, our algorithm exhibits superior PAC classification accuracy compared to the other methods.
\section{Conclusion}\label{sec-con}
This paper introduces a graph convolution network based on domain knowledge for ECG recognition, incorporating the landmarks points of PRQST as the ECG's domain knowledge. A temporal-spatial graph model is employed to depict both short-term intra-cycle abnormalities and long-term inter-cycle abnormalities. The effectiveness of our approach is validated through experiments on a large real-world dataset. The introduction of domain knowledge can improve the detection performance, especially for rare categories.

Further investigation is required to construct graph data with more intricate connections. It is noteworthy that our method extends beyond ECG and can find broad applications in cyclic signals (such as Photo Plethysmo Graphy, PPG) across various healthcare domains. Similarly, periodic signals in diverse fields like communications and meteorology can also be analyzed using this method.

%

\bibliographystyle{3870}
\bibliography{3870}

\end{sloppypar}
\end{document}